\documentclass[11pt]{article}
\usepackage[hyperref]{ccl2025-en}
\usepackage{times}
\usepackage{url}
\usepackage{latexsym}
\usepackage{graphicx}
\usepackage{booktabs}
\usepackage{multirow}
\usepackage{makecell}
\usepackage{amsmath}
\usepackage{amssymb}
\usepackage{fancyhdr}
\usepackage[table]{xcolor}
\usepackage{listings}
\usepackage{tcolorbox}
\tcbuselibrary{breakable,skins,listings}
\usepackage{enumitem}
\usepackage{tikz}
\usetikzlibrary{arrows.meta,positioning,fit,backgrounds,calc,shapes.geometric}

\definecolor{ourrow}{HTML}{DAE8FC}
\definecolor{baserow}{HTML}{EFF7D7}
\definecolor{midrow}{HTML}{F7E8DA}

\newtcolorbox{promptbox}[1]{
    breakable,
    enhanced,
    colback=cyan!3!white,
    colframe=cyan!75!black,
    coltitle=white,
    fonttitle=\bfseries\small,
    title=#1,
    boxrule=0.7pt,
    left=4pt, right=4pt, top=3pt, bottom=3pt,
    overlay middle={\draw[cyan!75!black, line width=0.7pt](frame.south west)--(frame.south east);},
    overlay last={\draw[cyan!75!black, line width=0.7pt](frame.south west)--(frame.south east);},
}
\definecolor{hlblue}{HTML}{1565C0}
\definecolor{hlgreen}{HTML}{2E7D32}
\definecolor{hlorange}{HTML}{E65100}
\definecolor{hlpurple}{HTML}{6A1B9A}
\definecolor{hlgray}{HTML}{616161}
\newcommand{\pkey}[1]{\textcolor{hlblue}{\textbf{#1}}}          
\newcommand{\ptype}[1]{\textcolor{hlgreen}{\texttt{#1}}}        
\newcommand{\prule}[1]{\textcolor{hlorange}{\textbf{#1}}}       
\newcommand{\pdesc}[1]{\textcolor{hlgray}{\textit{#1}}}         
\newcommand{\parrow}{$\rightarrow$}

\newcommand{\best}[1]{\textbf{#1}}
\newcommand{\secondbest}[1]{\underline{#1}}

\title{Retrieval-Augmented Multi-Prompt Ensemble for Minor-Grain Breeding Information Extraction}

\author{
  Hang Zhao$^{1}$\quad Jiahao Wang$^{2}$\thanks{Corresponding author. Email: wjh123king@gmail.com} \\
  $^{1}$Zhengzhou University \quad $^{2}$Harbin Institute of Technology (Shenzhen)
}

\date{}

\begin{document}
\maketitle

\begin{abstract}
This paper presents our system for CCL2026-Eval Task~5: Minor-Grain Breeding Information Extraction (MGBIE), 
which jointly extracts 12 entity types and 6 relation types from minor-grain breeding literature. 
We propose \textbf{RAME} (\textbf{R}etrieval-\textbf{A}ugmented \textbf{M}ulti-prompt \textbf{E}nsemble), 
a training-free framework that elicits multiple LLM outputs under controlled diversity and aggregates them by majority voting to obtain high-confidence predictions. 
RAME combines 
(i) retrieval-augmented few-shot selection via a hybrid BM25--embedding retriever, 
(ii) a three-prompt ensemble (\emph{Strict}, \emph{Relaxed}, \emph{Balanced}) spanning the precision to recall spectrum, 
and (iii) large-scale repeated sampling with majority voting to filter noisy predictions. 
Built on DeepSeek-V4-Flash, 
RAME achieves a Total Score of \textbf{0.499} (NER \textbf{0.730}, RE \textbf{0.346}) on the leaderboard, 
ranking \textbf{1st} and surpassing the official Track-A baseline powered by GPT-5.5 (0.448), representing an 11.4\% relative improvement. 
Code is available at \url{https://github.com/king-wang123/CCL26-RAME}.
\end{abstract}

\cclfootnote{
    \textcopyright 2026 China National Conference on Computational Linguistics

    \noindent This work is licensed under a Creative Commons Attribution 4.0 International License. License details: \url{http://creativecommons.org/licenses/by/4.0/}.
}

\setcounter{footnote}{0}

\section{Introduction}
\label{sec:intro}

Knowledge about minor-grain crops (e.g., millet, sorghum, buckwheat) is largely recorded in unstructured texts such as research papers and variety registration documents.
Such text is dense with domain terminology, exhibits variable surface forms for the same concept, and contains nested material names, 
all of which make key information hard to extract and structure consistently~\cite{lu2022uie}.

CCL2026-Eval Task~5 (MGBIE)\footnote{\url{https://github.com/zhiweihu1103/CCL2026-MGBIE}} addresses this challenge through two sub-tasks: 
\emph{Named Entity Recognition} (NER) over 12 fine-grained breeding entity types, 
and \emph{Relation Extraction} (RE) of 6 semantic relation types over the recognized entities. 
The task is difficult for three reasons: 
(i) it is highly domain-specialized with scarce annotated data; 
(ii) entity boundaries are fuzzy, since variety names, genes, QTLs, and markers take highly variable surface forms; 
and (iii) relations are diverse and often span multiple sentences.

Large language models (LLMs) have demonstrated strong capabilities across diverse NLP tasks~\cite{wang2026suco,brown2020language}. With in-context learning (ICL), they can adapt to specialized domains without parameter updates, making them particularly attractive for low-resource, schema-rich extraction settings. However, a single LLM decode is far from sufficient: the model is sensitive to prompt phrasing and demonstration selection, easily hallucinating spurious entities or missing valid relations~\cite{wang2025gptner}.
The question is how to reliably extract high-quality structured outputs from the model.

To this end, we propose \textbf{RAME} (\textbf{R}etrieval-\textbf{A}ugmented \textbf{M}ulti-prompt \textbf{E}nsemble), 
which performs repeated LLM inference under controlled diversity and selects high-consensus predictions. 
RAME consists of three components: 
(1) \emph{retrieval-augmented few-shot selection}, which retrieves topically similar training documents via a hybrid BM25--embedding retriever 
and randomly sub-samples $K$ demonstrations per draw to diversify the context; 
(2) a \emph{three-prompt ensemble} comprising \emph{Strict}, \emph{Relaxed}, and \emph{Balanced} variants 
that capture complementary extraction behaviours; 
and (3) \emph{large-scale repeated sampling with majority voting}, which aggregates $3\times N$ draws per document 
and retains only entities/relations recurring above a threshold. 
Voting acts as a test-time scaling mechanism~\cite{wang2023selfconsistency,snell2025scaling}: 
more draws improve recall, while a higher threshold improves precision.

Our contributions are as follows:
\begin{itemize}[leftmargin=1.4em,itemsep=2pt,topsep=2pt]
    \item We propose RAME, a fully training-free framework that integrates hybrid retrieval, a multi-prompt ensemble, and majority voting for domain information extraction, ranking \textbf{1st} on the leaderboard.
    \item We provide detailed ablations and analyze how voting threshold and sampling budget control the precision--recall trade-off.
    \item We release our code at \url{https://github.com/king-wang123/CCL26-RAME} to facilitate reproducibility.
\end{itemize}

\section{Related Work}
\label{sec:related}

\paragraph{LLMs for Information Extraction.}
Information extraction has shifted from task-specific tagging models toward generative formulations that directly emit structured outputs. 
Unified frameworks such as UIE~\cite{lu2022uie} and InstructUIE~\cite{wang2023instructuie} cast NER and RE as text-to-structure generation, 
while LLM-based methods like GPT-NER~\cite{wang2025gptner} and GoLLIE~\cite{sainz2024gollie} 
show that prompting with annotation guidelines enables competitive extraction without per-task supervised training. 
These approaches suit low-resource, domain-specific settings like MGBIE, but a single LLM decode often suffers from format drift and limited precision. 

\paragraph{Self-Consistency and Test-Time Scaling.}
Self-consistency~\cite{wang2023selfconsistency} samples multiple outputs and selects the most frequent answer by majority vote.
More broadly, allocating additional inference-time compute can be more effective than scaling model parameters~\cite{snell2025scaling,chen2025parscale}. Recent work such as SRAG-MAV~\cite{wang2025sragmav} demonstrates that multi-round inference with accumulative voting improves output stability and performance on structured prediction tasks.
While self-consistency votes over a single scalar answer, MGBIE outputs a \emph{set} of spans and triples. 
\section{Task Definition}
\label{sec:task}

\paragraph{Inputs and outputs.}
Given a passage $x$ of breeding text, a system must produce:
\begin{itemize}[leftmargin=1.4em,itemsep=1pt,topsep=2pt]
    \item An entity set $E$, where each entity is a character span annotated with one of 12 types: CROP, VAR, TRT, GST, GENE, QTL, MRK, CHR, BM, CROSS, ABS, BIS.
    \item A relation set $R$, where each relation is a triple $(h, t, r)$ linking a head entity $h$ and a tail entity $t$ with one of 6 relation types: CON (\emph{contains}), USE (\emph{uses}), HAS (\emph{has}), AFF (\emph{affects}), OCI (\emph{occurs\_in}), LOI (\emph{located\_in}).
\end{itemize}
Crucially, the relation taxonomy is coarse-grained: each relation type has typical head/tail entity type pairs (e.g., CROP$\rightarrow$VAR for CON), 
but valid pairs are \emph{not} limited to them, making RE substantially harder.

\paragraph{Evaluation.}
Both subtasks use exact matching. An NER true positive requires both boundary and type to be correct; an RE true positive requires the full triple $(h,t,r)$ to match. The scores are:
\begin{equation}
\small
\text{Score}_{\tau}=0.5\,F_1+0.25\,P+0.25\,R,\quad \tau\in\{\text{NER},\text{RE}\},
\end{equation}
\begin{equation}
\small
\text{Total}=0.4\,\text{Score}_{\text{NER}}+0.6\,\text{Score}_{\text{RE}}.
\end{equation}
The 0.6 weight on RE gives relation extraction greater weight in the final score.

\paragraph{Data.}   
We treat the provided 1{,}000 training documents as the retrieval pool. The 1{,}000 test documents are split into Phase-A (400) and Phase-B (600), with Phase-B serving as the final leaderboard set. Relations are sparse, averaging only 2.35 per document. 
\section{Method}
\label{sec:method}

\subsection{Overview}
\begin{figure}[t]
    \centering
    \includegraphics[width=\columnwidth]{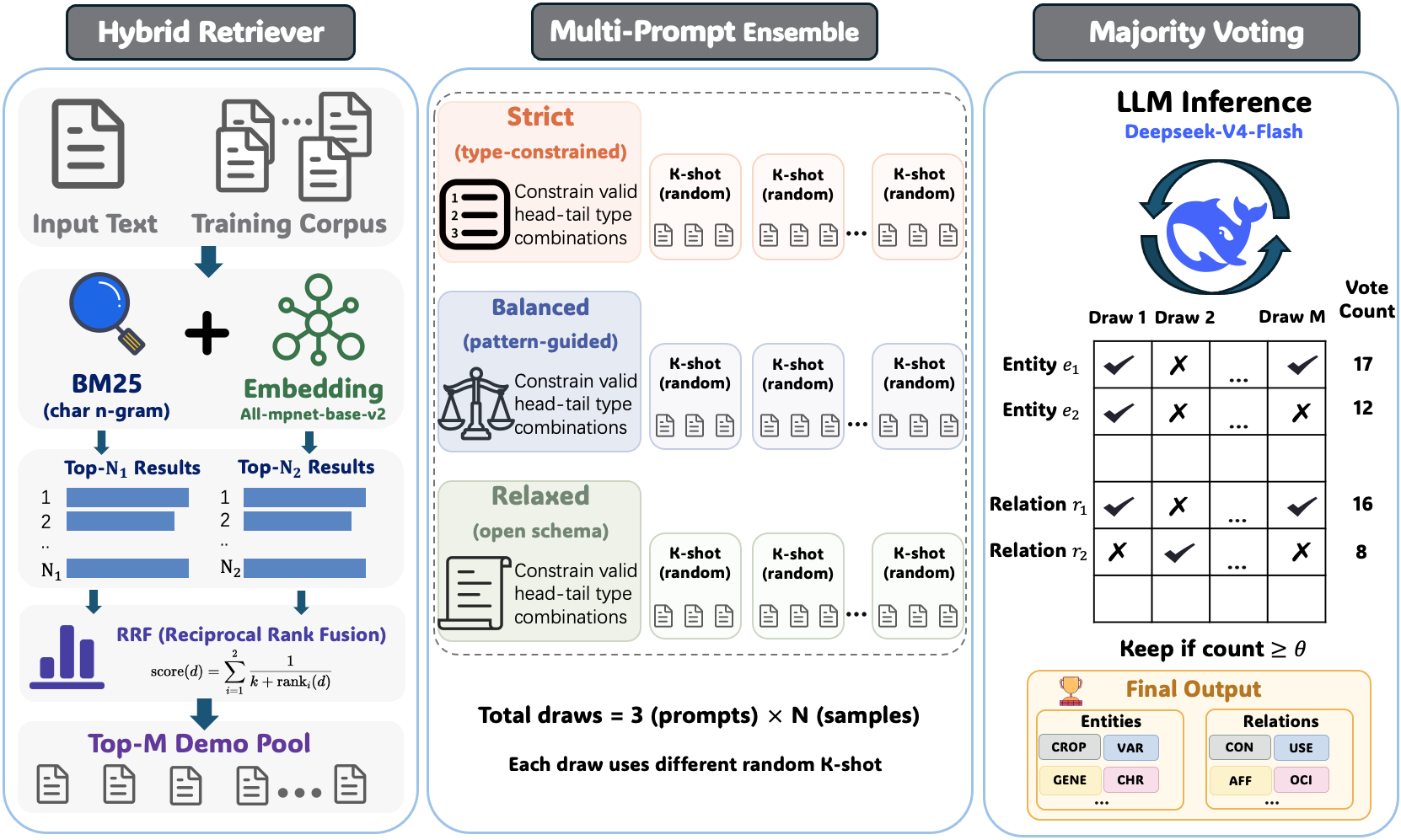}
    \caption{The RAME pipeline. A hybrid BM25--embedding retriever builds a candidate demonstration pool; three prompt variants (Strict/Relaxed/Balanced) each run $N$ draws with randomly sub-sampled $K$-shot demonstrations; outputs are aggregated by threshold voting over $3N$ draws.}
    \label{fig:pipeline}
\end{figure}

RAME is a training-free pipeline that samples multiple outputs from a single LLM under controlled diversity and aggregates them by voting~\cite{wang2025sragmav}. 
As shown in Figure~\ref{fig:pipeline}, for each document we: 
(1) retrieve a candidate pool of similar training documents with a hybrid retriever; 
(2) call the LLM $3N$ times using three prompt variants, each with $N$ independent draws and randomly sub-sampled $K$ demonstrations; 
(3) aggregate the $3N$ raw outputs by majority voting; and 
(4) align only the surviving entities and relations to character spans. 
We inject diversity at two levels: prompt variants and randomly sampled demonstrations. This makes errors independent, so they cancel out in the vote while correct predictions recur and survive.

\subsection{Retrieval-Augmented Few-Shot Selection}
\label{sec:method:retrieval}
Demonstration quality strongly affects ICL performance~\cite{liu2022makesgood}. Following the retrieval-augmented paradigm~\cite{lewis2020rag}, we combine two complementary retrievers over the training pool to select the most relevant examples:

\noindent\textbf{Lexical (BM25).} We tokenize text into character $n$-grams ($n\!\in\![3,6]$) and score with BM25~\cite{robertson2009bm25} ($k_1{=}1.2$, $b{=}0.75$). Character $n$-grams handle out-of-vocabulary identifiers (e.g., \texttt{qPH3.1}, \texttt{SSR-Xgwm20}) for which word-level matching is brittle.

\noindent\textbf{Semantic (embedding).} We encode documents with \texttt{all-mpnet-base-v2}~\cite{song2020mpnet,reimers2019sbert} (mean pooling, $L_2$-normalized, 768-d) and rank by cosine similarity.

\noindent\textbf{Hybrid fusion.} BM25 excels at surface-level matching while embedding captures semantic similarity; we fuse the two rankings via Reciprocal Rank Fusion (RRF)~\cite{cormack2009rrf} to benefit from both:
\begin{equation}
\small
\text{RRF}(d)=\sum_{m\in\{\text{bm25},\text{emb}\}}\frac{1}{k+\mathrm{rank}_m(d)},\quad k{=}60,
\end{equation}
and the top-$M$ documents form the candidate pool. To diversify draws, each draw \emph{randomly samples} $K$ demonstrations from this pool (with $M\!>\!K$), so different draws see different yet relevant contexts.

\subsection{Multi-Prompt Ensemble}
\label{sec:method:prompts}
A single instruction favours a particular extraction behaviour. An instruction that is too strict misses valid non-typical relations, whereas one that is too permissive over-generates. We therefore design three system prompts that share the same schema and output format but differ in their relation guidance:
\begin{itemize}[leftmargin=1.4em,itemsep=2pt,topsep=2pt]
    \item \textbf{Strict} emphasizes the typical typed head/tail pairs for each relation, favouring \emph{high precision}.
    \item \textbf{Relaxed} provides only the semantic definition and explicitly allows any head/tail combination, favouring \emph{high recall}.
    \item \textbf{Balanced} lists common patterns but states they are non-exhaustive, striking a middle ground.
\end{itemize}
Each variant excels at a different aspect of extraction; aggregating them combines their complementary strengths. The full prompts are given in Appendix~\ref{app:prompts}.

\subsection{Majority Voting over Structured Outputs}
\label{sec:method:voting}
Unlike self-consistency over a single answer~\cite{wang2023selfconsistency}, MGBIE outputs sets of entities and relational triples. We perform voting directly on the raw occurrence-based outputs, \emph{before} span alignment. An entity is keyed by its surface text, label, and occurrence index $(t,\ell,o)$; a relation by the full tuple $(h_t,h_\ell,h_o,t_t,t_\ell,t_o,r)$. We count each candidate at most once per draw and keep it if and only if it appears in at least $\theta$ of the $D{=}3N$ draws:
\begin{equation}
\small
\hat{Y}=\{\,c : \mathrm{votes}(c)\ge\theta\,\},\qquad \theta\in[1,D].
\end{equation}
The threshold $\theta$ controls the precision--recall trade-off: a small $\theta$ retains rare-but-correct predictions (higher recall), while a large $\theta$ keeps only high-consensus ones (higher precision). We use the strict-majority threshold $\theta{=}\lfloor D/2\rfloor+1$, which lies in the near-optimal region for the Total Score (Section~\ref{sec:exp}).

\subsection{Occurrence-Based Entity Alignment}
\label{sec:method:align}
Requiring the LLM to emit exact character offsets is unreliable. Instead, each predicted entity is referenced by its surface string, type, and an \emph{occurrence index} (the $k$-th appearance in the text). After voting selects the final set, we resolve each surviving entity to a character span via a cascade of increasingly tolerant matchers: exact substring, case-insensitive, whitespace-normalized, and punctuation-stripped. Relations reuse the resolved spans of their head/tail entities. Deferring alignment to after voting keeps the pipeline simple and avoids resolving spans for candidates that will be discarded.

\section{Experiments}
\label{sec:exp}

\subsection{Experimental Setup}
\paragraph{Datasets.}
For the official leaderboard, we use all 1{,}000 training documents as the retrieval pool and submit predictions on the 600-document Phase-B test set.
Since test labels are withheld, all ablations are conducted on a held-out split: 100 training documents serve as the development set and the remaining 900 as the retrieval pool.

\paragraph{Backbone and configuration.}
RAME performs no parameter tuning (Track-A).
The backbone is \textbf{DeepSeek-V4-Flash} deployed on $8\times$NVIDIA H20 (96\,GB) GPUs via sglang~\cite{zheng2024sglang}.
We use greedy decoding (temperature $0$), a maximum token budget of 65{,}536, and enable the model's thinking mode to elicit chain-of-thought reasoning~\cite{wei2022cot}, which improves extraction quality on complex schema-rich inputs.
The default configuration for all experiments is: $N{=}60$ draws per prompt ($3N{=}180$ per document),
voting threshold $\theta{=}91$ ($\approx 0.5\cdot 3N$), $K{=}8$ demonstrations from the top $M{=}30$ retrieved candidates,
and RRF fusion ($k{=}60$) of BM25 and \texttt{all-mpnet-base-v2}.

\begin{table}[t]
\centering
\small
\renewcommand{\arraystretch}{1.12}
\caption{Track-A (non-fine-tuning) leaderboard results. The official baseline uses GPT-5.5 with Prompt Engineering. \best{Bold} = best per column, \secondbest{underline} = second best. Our system (RAME) ranks 1st overall; RE remains the limiting factor.}
\label{tab:main}
\begin{tabular}{lccc}
\toprule
\textbf{System} & \textbf{NER} & \textbf{RE} & \textbf{Total} \\
 & \textbf{Score} & \textbf{Score} & \textbf{Score} \\
\midrule
\rowcolor{ourrow}
\textbf{RAME (Ours, 1st)} & \secondbest{0.730} & \secondbest{0.346} & \best{0.499} \\
2nd place & \best{0.736} & 0.337 & \secondbest{0.497} \\
3rd place & 0.715 & \best{0.350} & 0.496 \\
4th place & 0.725 & 0.337 & 0.492 \\
5th place & 0.725 & 0.334 & 0.491 \\
6th place & 0.719 & 0.335 & 0.488 \\
\midrule
\rowcolor{baserow}
Official baseline (GPT-5.5) & -- & -- & 0.448 \\
\bottomrule
\end{tabular}
\end{table}

\subsection{Main Results}
Table~\ref{tab:main} reports the Track-A leaderboard. 
RAME ranks \textbf{1st} with a Total Score of \textbf{0.499} without parameter tuning, 
exceeding the official Track-A Prompt-Engineering baseline powered by GPT-5.5 (0.448), representing an \textbf{11.4\% relative improvement}.
Its consistently strong performance across both NER (0.730) and RE (0.346), rather than a single dominant subtask score, yields the highest weighted Total Score.

\begin{table}[t]
\centering
\small
\renewcommand{\arraystretch}{1.12}
\caption{Ablation study on the 100-document held-out split. ``P$_i$'' denotes using only Prompt $i$ (Strict / Relaxed / Balanced) with a single draw; ``Ensemble'' merges all three prompts (single draw each); ``+Few-shot'' adds retrieval-augmented demonstrations; ``RAME'' further adds repeated sampling ($N{=}60$) with majority voting. Best in \best{bold}.}
\label{tab:ablation}
\begin{tabular}{lccc}
\toprule
\textbf{Configuration} & \textbf{NER} & \textbf{RE} & \textbf{Total} \\
\midrule
\rowcolor{baserow}
P$_1$ (Strict, zero-shot)   & 0.710 & 0.238 & 0.427 \\
\rowcolor{baserow}
P$_2$ (Relaxed, zero-shot)  & 0.686 & 0.186 & 0.386 \\
\rowcolor{baserow}
P$_3$ (Balanced, zero-shot) & 0.698 & 0.193 & 0.395 \\
\midrule
\rowcolor{midrow}
Ensemble (P$_{1{+}2{+}3}$, zero-shot) & 0.718 & 0.251 & 0.438 \\
\rowcolor{midrow}
+ Few-shot (RAG, single draw)          & 0.735 & 0.310 & 0.480 \\
\midrule
\rowcolor{ourrow}
\textbf{RAME} (+ repeated sampling \& voting) & \best{0.746} & \best{0.370} & \best{0.521} \\
\bottomrule
\end{tabular}
\end{table}

\subsection{Ablation Study}
Table~\ref{tab:ablation} isolates each RAME component. Rows~1--3 show the three individual prompts under zero-shot, single-draw conditions: Strict (P$_1$) achieves the highest single-prompt Total (0.427) owing to its strong RE precision, while Relaxed (P$_2$) trades NER precision for slightly higher RE recall. Ensembling all three (row~4) lifts Total to 0.438, a +0.011 gain over the best single prompt, confirming that their complementary tendencies benefit from simple output merging.

Adding retrieval-augmented demonstrations (row~5) produces the largest single jump (+0.042 Total), confirming that in-domain examples are essential for this specialized schema: without them the model lacks surface conventions for breeding identifiers such as gene names, QTLs, and marker codes. Repeated sampling with majority voting (RAME, row~6) adds a further +0.041, with RE benefiting most (+0.060). Voting suppresses spurious, low-consensus relations that a single decode tends to hallucinate.

\begin{figure}[t]
    \centering
    \includegraphics[width=0.92\columnwidth]{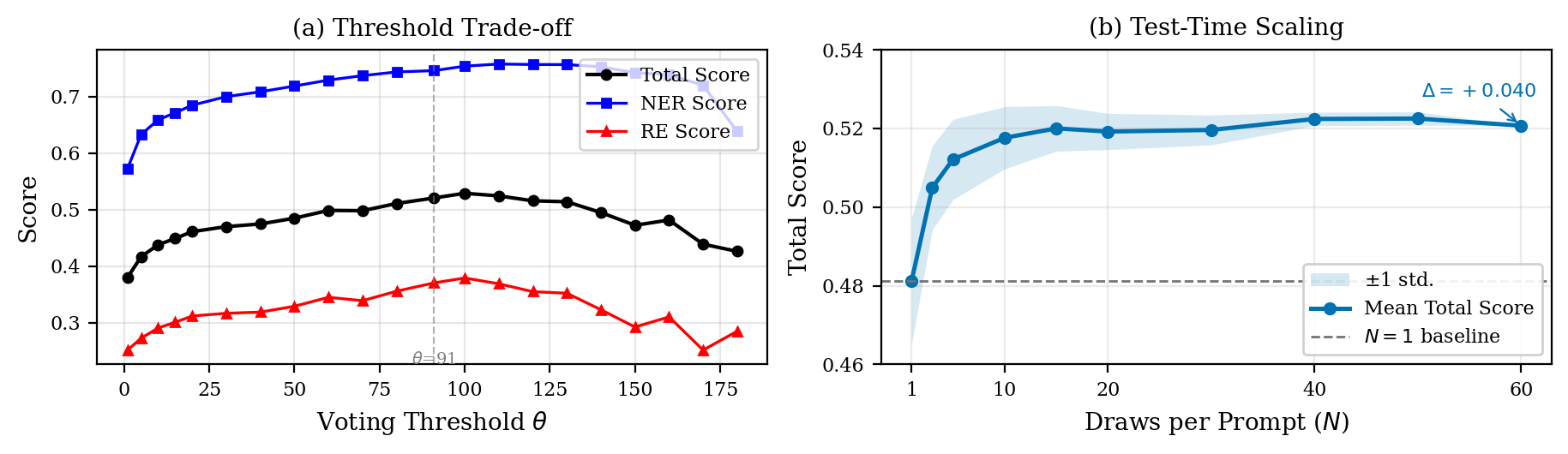}
    \caption{(a) NER, RE, and Total Scores under different voting thresholds. Total peaks at $\theta{=}100$, while the selected strict-majority threshold $\theta{=}91$ lies in the near-optimal region. (b) Mean Total Score under increasing test-time sampling budgets, with gains saturating around $N{=}15$. Each point ($N{<}60$) is averaged over 30 random subsets of the available draws; shading denotes $\pm$1 std., and $N{=}60$ uses all 180 draws.}
    \label{fig:curves}
\end{figure}

\subsection{Analysis}
\paragraph{Threshold and sampling budget.}
Figure~\ref{fig:curves}(a) shows a non-monotonic effect of $\theta$: NER, RE, and Total Scores initially rise as low-consensus predictions are filtered, but eventually decline as the criterion becomes too restrictive. The Total Score peaks at $\theta{=}100$ (0.529), while the selected strict-majority threshold $\theta{=}91$ attains 0.521 and lies in the same near-optimal region. For an order-insensitive estimate of test-time scaling (Figure~\ref{fig:curves}(b)), we randomly sample $N$ outputs from each prompt 30 times and aggregate with a strict majority rule $\theta(N){=}\lfloor 3N/2\rfloor+1$. Mean Total Score rises sharply from 0.481 at $N{=}1$ to 0.520 at $N{=}15$, but remains within 0.519--0.523 for $N{=}15$--60; increasing the budget from 45 to 180 total draws yields only a further 0.001. Test-time scaling therefore exhibits clear diminishing returns and effectively saturates around $N{=}15$, rather than improving linearly with inference compute~\cite{snell2025scaling}. Meanwhile, the standard deviation across random subsets contracts from 0.016 at $N{=}1$ to 0.006 at $N{=}15$ and 0.002 at $N{=}40$--50, indicating that additional draws primarily improve stability after performance saturates.

\section{Conclusion}
\label{sec:conclusion}

We presented RAME, our 1st-place system for CCL2026-Eval Task 5 (MGBIE). It combines hybrid retrieval-augmented few-shot selection, a three-prompt ensemble spanning the precision--recall spectrum, and large-scale repeated sampling with majority voting over occurrence-based outputs followed by span alignment. On the Track-A leaderboard, it achieves a Total Score of 0.499 without any parameter tuning, an 11.4\% relative improvement over the official GPT-5.5-based baseline. Voting turns single-pass LLM instability into a high-precision ensemble and serves as a controllable test-time scaling knob.

\paragraph{Limitations and future work.}
RAME trades inference compute for accuracy ($3N$ decodes per document) and is bounded by relation recall under exact-triple matching. Future directions include relation-centric and cross-sentence prompting, adaptive sampling with early stopping once votes stabilize, and distilling the ensemble into a single cheaper model. The code is released at \url{https://github.com/king-wang123/CCL26-RAME}.

\bibliography{references}

\newpage
\appendix
\section{Prompt Templates}
\label{app:prompts}

All three variants share the same system-prompt skeleton (entity definitions and output schema) and differ only in their \emph{relation guidance} block and the final \emph{Rules} line. We reproduce the shared skeleton once and then the three variant-specific blocks. Color coding: \pkey{key instructions} in blue bold, \ptype{type codes} in green monospace, \prule{rule emphasis} in orange bold, and \pdesc{descriptions} in gray italic.

\begin{promptbox}{Shared Skeleton -- Entity Guide + Output Schema}
\small
\pkey{You are an information extraction system for minor-grain crop breeding texts.}

\medskip
\textbf{Entities (12 types):}\\[2pt]
\begin{tabular}{@{}l@{\ }l@{}}
\ptype{CROP}  & \pdesc{Grain/cereal crop species or category} \\
\ptype{VAR}   & \pdesc{Named cultivar / line / accession / germplasm} \\
\ptype{TRT}   & \pdesc{Phenotype / agronomic / quality / resistance trait} \\
\ptype{GST}   & \pdesc{Developmental stage / phenology / time point} \\
\ptype{GENE}  & \pdesc{Gene or candidate gene name} \\
\ptype{QTL}   & \pdesc{Quantitative trait locus name/region} \\
\ptype{MRK}   & \pdesc{SSR/SNP/AFLP/Indel/KASP marker} \\
\ptype{CHR}   & \pdesc{Chromosome / linkage group / interval identifier} \\
\ptype{BM}    & \pdesc{Breeding / selection / experimental methodology} \\
\ptype{CROSS} & \pdesc{Parent material or cross / mapping population} \\
\ptype{ABS}   & \pdesc{Abiotic stress or treatment} \\
\ptype{BIS}   & \pdesc{Biological stress (pathogen / pest / disease)} \\
\end{tabular}

\medskip
\textbf{Relations (6 types):}\\
{\footnotesize\texttt{<<RELATION GUIDANCE -- variant-specific, see below>>}}

\medskip
\textbf{Rules:}\\
{\footnotesize\texttt{<<RULES -- variant-specific, see below>>}}

\medskip
\prule{Return ONLY:}
\begin{verbatim}
{
  "entities": [{"text": "<exact substring>",
                "label": "<CROP|VAR|...|BIS>",
                "occurrence": <int>}],
  "relations": [{"head": "<text>", "head_label": "<lbl>",
                 "head_occurrence": <int>,
                 "tail": "<text>", "tail_label": "<lbl>",
                 "tail_occurrence": <int>,
                 "label": "<CON|USE|HAS|AFF|OCI|LOI>"}]
}
\end{verbatim}
\end{promptbox}

\begin{promptbox}{Strict Variant \normalfont\pdesc{(targets high precision)}}
\small
\textbf{Relation guidance} --- \pkey{typed-pair constraints enforced}:

\medskip
\begin{tabular}{@{}l@{\ \ }l@{\ \ }l@{}}
\ptype{CON} & \pkey{typical:} \ptype{CROP}\parrow\ptype{VAR} & \pdesc{Variety belongs to a crop} \\
\ptype{USE} & \pkey{typical:} \ptype{VAR}\parrow\ptype{BM}   & \pdesc{Breeding method used to produce a variety} \\
\ptype{HAS} & \pkey{typical:} \ptype{VAR}\parrow\ptype{TRT}  & \pdesc{Variety has / exhibits a trait} \\
\ptype{AFF} & \pkey{typical:} \ptype{ABS/GENE/MRK/QTL}\parrow\ptype{TRT} & \pdesc{Entity affects a trait} \\
\ptype{OCI} & \pkey{typical:} \ptype{TRT/ABS/BIS}\parrow\ptype{GST}      & \pdesc{Trait/stress at a growth stage} \\
\ptype{LOI} & \pkey{typical:} \ptype{MRK/QTL/GENE}\parrow\ptype{CHR}     & \pdesc{Entity located on a chromosome} \\
\end{tabular}

\medskip
\textbf{Rules:} \prule{entities are verbatim substrings}; extract \pkey{every mention} (\texttt{occurrence=1,2,3\ldots}); \prule{relations must use extracted entities} as head/tail.
\end{promptbox}

\begin{promptbox}{Relaxed Variant \normalfont\pdesc{(targets high recall)}}
\small
\textbf{Relation guidance} --- \pkey{ANY head/tail type combination is valid} when the semantic definition fits:

\medskip
\begin{tabular}{@{}l@{\ \ }l@{}}
\ptype{CON} & \pdesc{Head contains/subsumes the tail} \\
\ptype{USE} & \pdesc{Head uses the tail as a method/marker/resource} \\
\ptype{HAS} & \pdesc{Head possesses/exhibits/is characterised by the tail trait} \\
\ptype{AFF} & \pdesc{Head affects/regulates/correlates with the tail} \\
\ptype{OCI} & \pdesc{Head occurs in/is measured at the tail growth stage} \\
\ptype{LOI} & \pdesc{Head is located in or associated with the tail} \\
\end{tabular}

\medskip
\textbf{Rules:} \prule{entities are verbatim substrings}; extract \pkey{every mention} (\texttt{occurrence=1,2,3\ldots}); \prule{err on the side of recall} for relations.
\end{promptbox}

\begin{promptbox}{Balanced Variant \normalfont\pdesc{(precision--recall trade-off)}}
\small
\textbf{Relation guidance} --- \pkey{common patterns shown but NOT exhaustive}; semantic definition is authoritative:

\medskip
\begin{tabular}{@{}l@{\ \ }l@{\ \ }l@{}}
\ptype{CON} & \pkey{common:} \ptype{CROP}\parrow\ptype{VAR} & \pdesc{Head contains/subsumes the tail. \prule{Other combos also valid.}} \\
\ptype{USE} & \pkey{common:} \ptype{VAR}\parrow\ptype{BM}   & \pdesc{Head uses the tail. \prule{Other combos also valid.}} \\
\ptype{HAS} & \pkey{common:} \ptype{VAR}\parrow\ptype{TRT}  & \pdesc{Head possesses/exhibits the tail. \prule{Other combos also valid.}} \\
\ptype{AFF} & \pkey{common:} \ptype{ABS/.../QTL}\parrow\ptype{TRT} & \pdesc{Head affects the tail. \prule{Other combos also valid.}} \\
\ptype{OCI} & \pkey{common:} \ptype{TRT/.../BIS}\parrow\ptype{GST} & \pdesc{Head occurs at the tail stage. \prule{Other combos also valid.}} \\
\ptype{LOI} & \pkey{common:} \ptype{MRK/.../GENE}\parrow\ptype{CHR} & \pdesc{Head located in/assoc.\ with the tail. \prule{Other combos also valid.}} \\
\end{tabular}

\medskip
\textbf{Rules:} \prule{entities are verbatim substrings}; extract \pkey{every mention} (\texttt{occurrence=1,2,3\ldots}); \pkey{prefer common pattern} when it applies but \prule{don't skip non-typical relations}.
\end{promptbox}

\medskip
Each call appends $K$ retrieved demonstrations (gold passages with their labeled JSON) followed by the target passage under the template:

\begin{promptbox}{User Message Template}
\small
\pkey{Example \{i\}:}\\
Passage:\\
\texttt{"""}\\
\texttt{\{demo\_text\}}\\
\texttt{"""}\\
Output:\\
\texttt{\{demo\_labeled\_json\}}

\medskip
\ldots\ ($K$ examples total)

\medskip
\pkey{Now the real passage:}\\
Passage:\\
\texttt{"""}\\
\texttt{\{text\}}\\
\texttt{"""}\\[2pt]
\prule{Return the JSON object.}
\end{promptbox}

\section{Implementation Details}
\label{app:impl}

\paragraph{Hyperparameters.}
Table~\ref{tab:hparams} lists the configuration used for all leaderboard submissions and (on the held-out split) for ablations.

\begin{table}[h]
\centering
\small
\renewcommand{\arraystretch}{1.1}
\caption{RAME hyperparameters.}
\label{tab:hparams}
\begin{tabular}{lll}
\toprule
\textbf{Symbol} & \textbf{Value} & \textbf{Meaning} \\
\midrule
$N$ & 60 & draws per prompt variant \\
$3N$ & 180 & total draws per document \\
$\theta$ & 91 & voting threshold ($\approx 0.5\cdot 3N$) \\
$K$ & 8 & demonstrations per draw \\
$M$ & 30 & retrieval candidate pool size \\
$k_{\text{RRF}}$ & 60 & RRF fusion constant \\
temperature & 0.0 & decoding temperature \\
max tokens & 65{,}536 & reasoning + answer budget \\
BM25 $(k_1,b)$ & $(1.2,0.75)$ & char $n$-gram ($n\!\in\![3,6]$) \\
\bottomrule
\end{tabular}
\end{table}

\paragraph{Backbone and serving.}
The backbone LLM is DeepSeek-V4-Flash, served with sglang behind an OpenAI-compatible \texttt{/chat/completions} API on $8\times$NVIDIA H20 (96\,GB) GPUs (driver 535.161.07, CUDA 12.8). Thinking/reasoning mode is enabled; because reasoning and the answer share the token budget, a draw whose content is empty due to length truncation is retried with a doubled budget.

\paragraph{Development split.}
For offline analysis we randomly partition the 1{,}000 training documents into a 900-document retrieval pool and a 100-document development set; the split indices are fixed across all experiments so that numbers are comparable.

\paragraph{Span alignment and JSON parsing.}
The model emits occurrence-based references rather than offsets. We resolve each $(\text{text},\text{label},\text{occurrence})$ to a span via the cascade: exact substring $\rightarrow$ case-insensitive $\rightarrow$ whitespace-normalized $\rightarrow$ punctuation-stripped; relations reuse their head/tail resolved spans, and a relation is dropped if either endpoint cannot be aligned. JSON is extracted robustly by stripping any \texttt{<think>} block and markdown fences and parsing the last balanced top-level object; malformed draws contribute nothing to the vote. Entities and relations are deduplicated by their span/type and full-tuple keys before voting and after aggregation.

\end{document}